%% file: paper.tex
\documentclass[10pt,twocolumn,letterpaper]{article}

\usepackage{cvpr}              % To produce the CAMERA-READY version
\input{preamble}
\definecolor{cvprblue}{rgb}{0.21,0.49,0.74}
\usepackage[pagebackref,breaklinks,colorlinks,allcolors=cvprblue]{hyperref}
\usepackage{multicol}
\usepackage{booktabs}
\usepackage{siunitx}
\usepackage{censor}
\usepackage{xcolor}
\usepackage{multirow}
\usepackage{layouts}
\usepackage{placeins}
\usepackage{cuted}

\def\paperID{17625} % *** Enter the Paper ID here
\def\confName{CVPR}
\def\confYear{2026}

\title{Another BRIXEL in the Wall: Towards Cheaper Dense Features}

\author{%
Alexander Lappe$^{1,2}$ \quad \quad Martin A. Giese$^{1}$
\\
$^1$Hertie Institute, University of Tübingen \quad $^2$IMPRS-IS \\
\texttt{alexander.lappe@uni-tuebingen.de}
}

\begin{document}
\maketitle

\begin{strip}
\centering
\includegraphics[width=\textwidth]{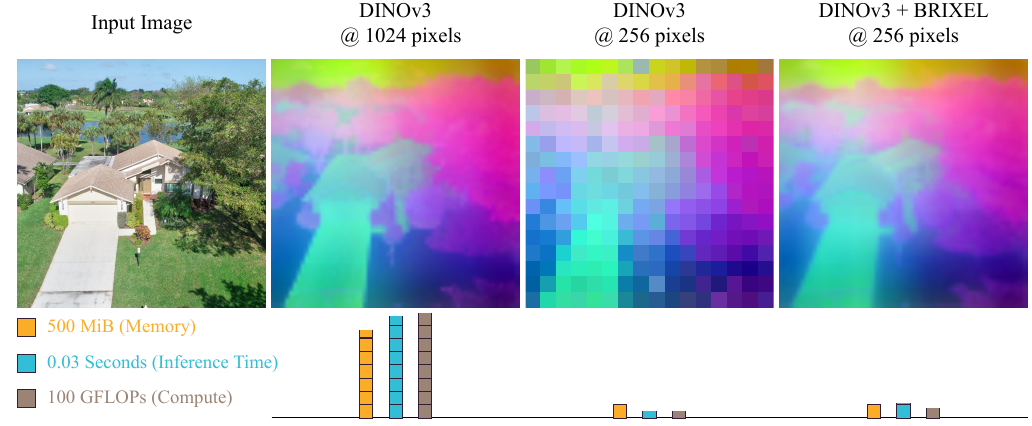}
\captionof{figure}{Recent dense feature extractors are able to operate at very high resolution, albeit at great computational cost. We propose BRIXEL, a simple self-distillation approach that produces dense feature maps while circumventing the Vision Transformer's quadratic scaling.}
\label{fig:overview}
\end{strip}

\begin{abstract}
  Vision foundation models achieve strong performance on both global and locally dense downstream tasks. Pretrained on large images, the recent DINOv3 model family is able to produce very fine-grained dense feature maps, enabling state-of-the-art performance. However, computing these feature maps requires the input image to be available at very high resolution, as well as large amounts of compute due to the squared complexity of the transformer architecture. To address these issues, we propose BRIXEL, a simple knowledge distillation approach that has the student learn to reproduce its own feature maps at higher resolution. Despite its simplicity, BRIXEL outperforms the baseline DINOv3 models by large margins on downstream tasks when the resolution is kept fixed. We also apply BRIXEL to other recent dense-feature extractors and show that it yields substantial performance gains across model families. Code and model weights are available at
% \censor{githubblablalba}.
https://github.com/alexanderlappe/BRIXEL.
\end{abstract}

\section{Introduction}
\label{sec:intro}

In the late 1970s, Pink Floyds's \textit{Another Brick in the Wall} raised the question of whether teachers actually make their students smarter. Here, we revisit the idea of a teacher that makes its students \textit{more} dense in the context of self-distillation of dense image features. Great strides have been made in unsupervised pretraining of vision foundation models in recent years. The highly flexible Vision Transformer (ViT) \cite{dosovitskiyImageWorth16x162021} architecture has enabled a variety of very powerful, general-purpose feature extractors \cite{radfordLearningTransferableVisual2021, zhaiSigmoidLossLanguage2023, zhouIBOTImageBERT2021, touvronDeiTIIIRevenge2022, caronEmergingPropertiesSelfSupervised2021a, oquabDINOv2LearningRobust2023, simeoniDINOv32025, bolyaPerceptionEncoderBest2025}. One of the most intriguing properties of these ViT-based models is that they learn not only global image representations, but also dense descriptors for local image regions. While the quality of these dense features varies across models \cite{simeoniDINOv32025, bananiProbing3DAwareness2024}, the state-of-the-art models for fine-grained spatial tasks such as depth estimation \cite{yangDepthAnythingV22024} or segmentation \cite{raviSAMSegmentAnything2024, simeoniDINOv32025} rely on this mechanism.

The main disadvantage of using Vision Transformers for dense tasks is that the spatial resolution of the features is inherently limited. Since these models divide the input into local patches and, unlike convolutional networks, operate at the same spatial resolution throughout the entire architecture, the final feature resolution is equal to the resolution of input image patches. The most commonly used patch size is 16, meaning that the feature maps will be downsampled by that factor relative to the input image. One way to circumvent this problem is to pretrain the model on very-high resolution images, which results in extremely high-resolved feature maps at test time \cite{simeoniDINOv32025}. However, this strategy comes with two caveats: First, the test-time or downstream task images need to be available in much higher resolution than the desired feature resolution, which in many applications will not be the case. Second, the quadratic scaling of the computational complexity of the transformer architecture with respect to the number of input tokens makes this approach very expensive. 

This disadvantage is usually addressed by decoupling the heavy semantic and geometric lifting of the ViT from the fine-grained spatial computations. The high-performing dense ViTs mentioned in the previous paragraph rely on a pretrained backbone, and a supervised, task-specific spatial refiner \cite{yangDepthAnythingV22024, raviSAMSegmentAnything2024, simeoniDINOv32025} like the ViT adapter \cite{chenVisionTransformerAdapter2022} or MaskFormer \cite{chengPixelClassificationNot2021, chengMaskedattentionMaskTransformer2022}. While leading to great performance, this approach requires substantial further supervised training after the unsupervised pretraining stage, meaning that it is not applicable when little supervised data is available. Therefore, it does not fully align with the desired off-the-shelf downstream task capabilities of these models.

\paragraph{Contribution.}

To remedy these issues, the goal of this work is to produce high-resolution, \emph{task-agnostic} dense features without any supervision. We summarize our contribution as follows: 
\begin{enumerate}
    \item Leveraging properties of the recently published DINOv3 model family, we propose BRIXEL, a simple yet powerful method to generate high-resolution dense features at a fraction of the computational cost.
    \item We evaluate the proposed method on a variety of dense downstream tasks. Across 42 model comparisons, our method outperforms the baseline at fixed resolution on every single one. Given the popularity of DINO models, and DINOv3's position as the state-of-the-art dense-feature model, these results promise performance gains and resource savings for practitioners in the field.
    \item We test the generalization abilities of BRIXEL and show that it yields substantial performance gains for other model families as well.
\end{enumerate}

\section{A teacher that makes the student more dense}
\begin{figure*}
    \centering
    \includegraphics[width=\linewidth]{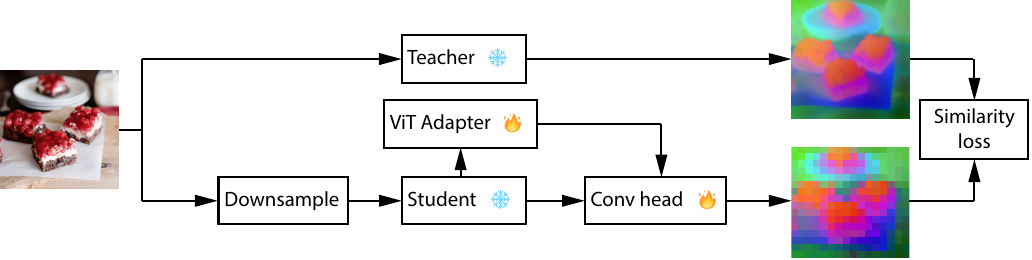}
    \caption{An overview of BRIXEL. The teacher and student network share both architecture and weights, which are all frozen. During training, the student receives a downsampled input image and has to reconstruct the dense features computed by the high-resolution teacher model. To do so, the student is connected to a standard ViT adapter and feeds into a convolutional readout head which fuses the output of the frozen student backbone and the trainable ViT adapter.}
    \label{fig:method}
\end{figure*}

As discussed in the previous section, prior work combined pretrained Vision Transformers with a fine-grained refiner module for dense tasks. While training the refiner, the transformer weights may be fine-tuned \cite{raviSAMSegmentAnything2024} as well or kept frozen \cite{simeoniDINOv32025}. In any case, optimizing the weights of the refiner network requires an additional learning signal beyond the (usually unsupervised) pretraining strategy, which is supplied in the form of task-specific label supervision \cite{raviSAMSegmentAnything2024, yangDepthAnythingV22024, simeoniDINOv32025}. 

Since we aim to produce task-agnostic high-resolution features, producing a strong learning signal for the refiner is not as straightforward. However, the recently published DINOv3 model family offers an interesting avenue in that direction. These models are trained at a variety of input resolutions, and are regularized to produce highly consistent dense features across image sizes \cite{simeoniDINOv32025}. As prior work has shown that the features of foundation models, coupled with a refiner network, perform well on a variety of tasks, we wonder whether the reconstruction of higher-resolution foundation model features is among those tasks.

\paragraph{Main idea.} We propose a simple method for generating high-resolution dense features by distilling fine-grained spatial information captured by a model at high resolution into a refiner network. The goal of the refiner is to operate in conjunction with the model at low resolution to output the same dense feature map that the model would produce for higher-resolution input. We sketch the training procedure in \cref{fig:method}. It consists of a simple teacher-student setup where the teacher and the student are identical networks with shared, frozen weights. The student is connected to a refiner network and both feed into a convolutional head that outputs the final dense features. During training, the teacher receives input images at high resolution, and the student receives the same image downsampled by a factor of 4 per side. The weights of the refiner network and the head are optimized for the output to mimic that of the teacher.

\paragraph{Aligning the feature maps.}
Let $\bf x \in \mathbb{R}^{3\times h\times w}$, and $\bf x_-\in \mathbb{R}^{3\times \frac{h}{4}\times \frac{w}{4}}$ denote an input image at high/low resolution respectively. Further, let $T(\cdot)$ denote the teacher and $S_{\boldsymbol{\theta}}(\cdot)$ the student network including the refiner and head, where $\boldsymbol{\theta}$ refers to the trainable parameters. First, we consider the $L_1$-loss between the outputs, i.e.
\begin{equation}
    \mathcal{L}_1(\boldsymbol{\theta}) := \mathbb{E}_{\mathbf{x}\sim p(\mathbf{x})}[||T(\mathbf{x}) - S_{\boldsymbol{\theta}}(\mathbf{x}_-)||_1].
\end{equation}
Empirically, we found that this loss function alone resulted in blurry boundaries. To ensure that the refiner produces faithful boundaries, we therefore also encourage it to match the output of Sobel edge detectors in feature space between the student's and teacher's feature maps. As edges at the single feature level are noisy, we compute an SVD on teacher tokens in each batch with gradients detached to find a projection $P$ onto the $K$ highest-variance principal components. Letting $\nabla_x$ and $\nabla_y$ denote the channel-wise Sobel operators, we define the edge loss
\begin{align}
    \mathcal{L}_{\text{edge}}(\boldsymbol{\theta}) := \mathbb{E}_{\mathbf{x}\sim p(\mathbf{x})}[||\nabla_x P(T(\mathbf{x})) - \nabla_x P(S_{\boldsymbol{\theta}}(\mathbf{x}_-))||_1 \\
    + ||\nabla_y P(T(\mathbf{x})) - \nabla_y P(S_{\boldsymbol{\theta}}(\mathbf{x}_-))||_1].
\end{align}
Here, the projection $P$ operates token-wise, i.e.
\begin{equation}
    P:\mathbb R^{C\times \tfrac{h}{p} \times \tfrac{w}{p}} \to \mathbb R^{C{\text{reduced}}\times \tfrac{h}{p} \times \tfrac{w}{p}},
\end{equation}
and $p$ denotes the patch size of the model.
Finally, we also include a spectral loss to encourage similar high-frequency components between student and teacher output. To this end, we compute the FFT of both feature maps, convert them to polar coordinates and average amplitudes over concentric circles with fixed radius $r$ to obtain one-dimensional frequency spectra $p_{T(\mathbf{x})}(r)$ and $p_{S(\mathbf{x}_-)}(r)$. We compare high-frequency spectra using the loss
\begin{align}
&\mathcal{L}_{\text{spectral}}(\boldsymbol{\theta}) \\
&= \mathbb{E}_{\mathbf{x}\sim p(\mathbf{x})}\!\Big[\nonumber
    \frac{1}{|\mathcal{R}|}
    \sum_{r \in \mathcal{R}}
    \big(\log p_{T(\mathbf{x})}(r) - \log p_{S(\mathbf{x}_-)}(r)\big)^2
\Big],
\end{align}
where $\mathcal{R} := \{ r |r \geq r_0 \}$ contains the high-frequency components.
Finally, the overall loss becomes
\begin{equation}
    \mathcal L_{\text{total}}(\boldsymbol{\theta}) := \mathcal L_1 (\boldsymbol{\theta}) + \lambda_{\text{edge}}\mathcal L_{\text{edge}}(\boldsymbol{\theta}) + \lambda_{\text{spectral}} \mathcal L_{\text{spectral}}(\boldsymbol{\theta}). 
\end{equation}

\section{Experiments}
\paragraph{Backbones.} We first perform experiments on the pretrained DINOv3 models published at https://github.com/facebookresearch/dinov3. To cover the range of available model sizes, we select the small, base, large, and huge+ variants of the ViT architecture, which differ only in the number of attention blocks and internal dimensions. To study generalization to other model families in \cref{subsec:beyond_dino}, we use the ViT-Base variants of SIGLIP 2 \cite{tschannenSigLIPMultilingualVisionLanguage2025} and Perception Encoder Spatial \cite{bolyaPerceptionEncoderBest2025} with the pretrained weights published on huggingface.co.

\paragraph{ViT Adapter.}
For the adapter network, we utilize the same model that was previously used to apply DINOv3 to supervised dense tasks \cite{simeoniDINOv32025}. It is based on the initial ViT adapter \cite{chenVisionTransformerAdapter2022}, with the only difference being that it does not feed back into the ViT backbone in which all weights are frozen. It  consists of a spatial prior module, implemented by a convolutional network operating on the image itself to extract spatial information, and a multi-scale feature extractor which extracts features at various resolutions from the ViT tokens. We utilize only the highest-resolution features output by the ViT adapter, which have $1/4$ of the input image resolution, and feed these into the final convolutional head, which consists of three depthwise-separable convolution layers. Finally, we obtain the high-resolution dense features by adding the output to a bilinearly upsampled version of the original ViT backbone features. We use the implementation of the ViT adapter provided in \cite{simeoniDINOv32025}, attach the convolutional head and then train the weights for our purpose from scratch.

\paragraph{Data.} As the training is self-supervised, the only necessary criterion for our training data is that its resolution needs to be sufficient for the teacher network to compute high-quality target features. For simplicity, we randomly sample high-resolution images from LAION and the Segment anything database to obtain a training set of 110k images. 

\paragraph{Training.} We then train the four adapter networks at a resolution of $256^2$ (so the teacher network operates at a resolution of $1024^2$). Each model is trained using Adam on a single NVIDIA A100 for a total of 40k iterations. We set $\lambda_\text{edge}=1$, $\lambda_\text{spectral}=0.1$, $K=8$ and the learning rate to $1\cdot 10^{-3}$ with one warmup epoch.

\subsection{Qualitative evaluation}
\begin{figure*}
    \centering
    \includegraphics[width=\textwidth]{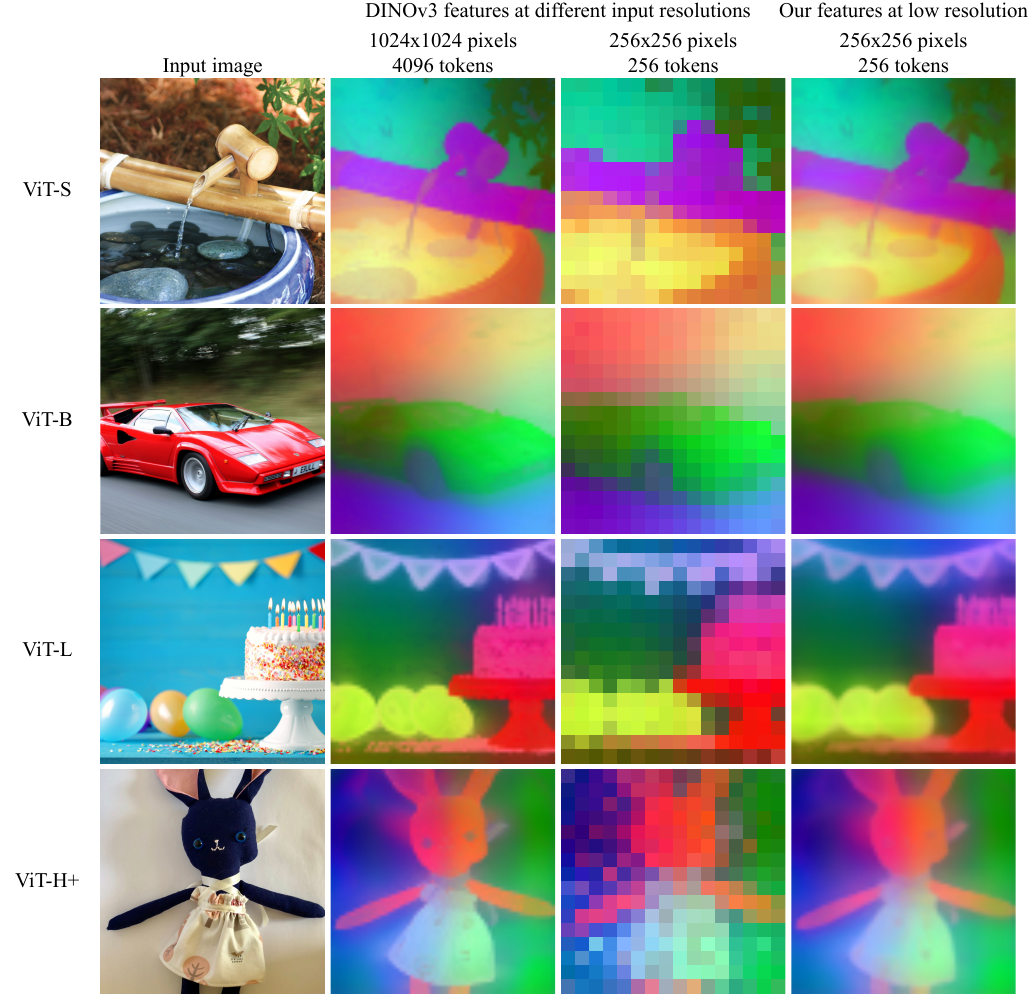}
    \caption{Qualitative evaluation of the proposed method. The second and third column display the dense feature maps of DINOv3 when feeding in the input image at different resolutions. At 256 pixels per side, feature maps become very blurry. The final column shows the dense feature maps of DINOv3 when combined with BRIXEL. Even though we also input images at 256 pixels per side, the feature maps are visually almost indistinguishable from those computed with 4096 tokens at considerably higher computational cost. As has become standard practice, we create the visualizations by performing a singular value decomposition on the 4096 tokens of the high-resolution target feature map. Then, we project all tokens of all images in the same row onto the first three singular vectors and map the results to RGB values.}
    \label{fig:qualitative_panel}
\end{figure*}

We show qualitative results in \cref{fig:qualitative_panel} by performing a principal component analysis on the teacher tokens and mapping all tokens to RGB. We observe that the low-resolution DINOv3 features allow the adapter to recover very fine-grained spatial representations. In some cases, the teacher output and the student output appear visually almost indistinguishable in PCA space. This is true across all model sizes we tested, and visually we did not observe differences in feature map quality between model sizes.

\subsection{Evaluation on scene-centric tasks}
\begin{table}[t]
\centering
\small
\setlength{\tabcolsep}{3.5pt}
\caption{Performance of lightweight probes on top of frozen backbone models for scene-centric tasks. 'Baseline' refers to the original DINOv3 model, 'Ours' refers to the student DINOv3 model equipped with a general-purpose ViT adapter. We report mean Intersection over Union and Pixel Accuracy for semantic segmentation and Root Mean Square Error for monocular depth estimation.}
\label{tab:scene-centric}

% ------------------ Segmentation (combined) ------------------
\begin{tabular}{l
                S[table-format=2.1] S[table-format=2.1] S[table-format=2.1] S[table-format=2.1]
                S[table-format=2.1] S[table-format=2.1] S[table-format=2.1] S[table-format=2.1]}
\toprule
& \multicolumn{4}{c}{ADE20k} & \multicolumn{4}{c}{Cityscapes} \\
\cmidrule(lr){2-5}\cmidrule(lr){6-9}
& \multicolumn{2}{c}{mIoU ↑} & \multicolumn{2}{c}{Pixel Acc. ↑}
& \multicolumn{2}{c}{mIoU ↑} & \multicolumn{2}{c}{Pixel Acc. ↑} \\
\cmidrule(lr){2-3}\cmidrule(lr){4-5}\cmidrule(lr){6-7}\cmidrule(lr){8-9}
\textbf{Model Size}
& {Base} & {Ours} & {Base} & {Ours}
& {Base} & {Ours} & {Base} & {Ours} \\
\midrule
Small & 41.4 & \bfseries 43.5 & 78.5 & \bfseries 80.0 & 56.7 & \bfseries 57.7 & 90.6 & \bfseries 91.6 \\
Base  & 46.7 & \bfseries 49.2 & 80.5 & \bfseries 82.0 & 61.1 & \bfseries 64.4 & 91.6 & \bfseries 93.0 \\
Large & 49.8 & \bfseries 52.5 & 81.2 & \bfseries 82.9 & 63.3 & \bfseries 66.7 & 91.9 & \bfseries 93.3 \\
Huge+ & 49.0 & \bfseries 52.1 & 80.5 & \bfseries 82.3 & 64.0 & \bfseries 68.0 & 92.0 & \bfseries 93.5 \\
\bottomrule
\end{tabular}

\vspace{6pt}

% ------------------ Depth (separate) ------------------
\begin{tabular}{l S[table-format=1.3] S[table-format=1.3]}
\toprule
\multicolumn{3}{c}{NYU (RMSE ↓)} \\
\cmidrule(lr){1-3}
\textbf{Size} & {Baseline} & {Ours} \\
\midrule
Small & 0.382 & \bfseries 0.376 \\
Base  & 0.354 & \bfseries 0.346 \\
Large & 0.335 & \bfseries 0.320 \\
Huge+ & 0.348 & \bfseries 0.332 \\
\bottomrule
\end{tabular}

\end{table}

Next, we probe whether the increased spatial resolution of the student's feature maps leads to performance gains on dense tasks over the baseline DINOv3 model. We start by considering scene-centric problems. In particular, we test the models on semantic scene segmentation using the benchmarks ADE20k \cite{zhouSceneParsingADE20K2017, zhouSemanticUnderstandingScenes2018} and Cityscapes \cite{cordtsCityscapesDatasetSemantic2016} at a resolution of 256, as well as monocular depth estimation on NYU \cite{silbermanIndoorSegmentationSupport2012} at $288\times 384$. For segmentation, we train a linear probe on top of the frozen backbone model. For depth estimation, we train a lightweight non-linear probe adapted from \cite{bananiProbing3DAwareness2024}, which again takes only the frozen last-layer features as input. As depth estimation is slightly more involved than classification, we follow the implementation details from \cite{bananiProbing3DAwareness2024} and kindly refer the reader to their work for details.
Importantly, the weights of the ViT adapter are frozen during all experiments beyond the self-distillation stage, as we explicitly aim to obtain more dense feature maps without relying on external supervision. As all feature maps have lower resolution than the input images and we are considering pixel-level tasks, we bilinearly interpolate the output of each probe to pixel-level resolution. We report mean Intersection over Union (mIoU) and Pixel Accuracy for semantic segmentation and root mean squared error for depth estimation in \cref{tab:scene-centric}. 

Across all models, as well as both segmentation datasets and both metrics, we observe substantial performance increases over the base DINOv3 model. These findings quantitatively corroborate the intuition gained from the visualizations in \cref{fig:qualitative_panel}, which show very high spatial precision of the student's feature maps. For scene-centric depth estimation, we see modest but consistent improvements over the baseline. As depth varies primarily at low frequencies and the best depth estimation systems also incorporate mid-level features, we hypothesize that the bottleneck for depth estimation lies with the features themselves rather than spatial precision. This could explain why performance gains are larger for segmentation tasks.

\subsection{Evaluation on object-centric tasks}
We also evaluate the student on fine-grained object-centric tasks. We perform object-part segmentation on PASCAL VOC \cite{chenDetectWhatYou2014}, and monocular depth estimation, as well as surface normal estimation on NAVI \cite{jampani2023navi}. For the former two tasks, we follow the same protocols as in the previous section and report the same metrics. For the object images from the NAVI dataset, we also compute surface normals from the depth maps and predict them using the same probe as for depth without sharing weights. For this task, we report the mean Angular Error in degrees. \cref{tab:object-centric} displays the results. Again, we see a substantial performance increase relative to the baseline DINOv3 model on the segmentation task, showing that the student's feature maps capture fine-grained object characteristics as well as scene semantics. We also consistently outperform the baseline on object-centric depth estimation and surface normal estimation.

\begin{table}[t]
\centering
\small
\setlength{\tabcolsep}{3.5pt}
\caption{Performance of lightweight probes on top of frozen backbone models for object-centric tasks. For object part segmentation on PASCAL VOC, we report the same metrics as for semantic scene segmentation in \cref{tab:scene-centric}. For the object-centric NAVI dataset, we report RMSE for depth estimation as well as the mean angle between predicted and ground-truth surface normal vectors.}
\label{tab:object-centric}

\begin{tabular}{l
                S[table-format=2.1] S[table-format=2.1] S[table-format=2.1] S[table-format=2.1]
                S[table-format=1.3] S[table-format=1.3] S[table-format=2.2] S[table-format=2.2]}
\toprule
& \multicolumn{4}{c}{PASCAL VOC} & \multicolumn{4}{c}{NAVI} \\
\cmidrule(lr){2-5}\cmidrule(lr){6-9}
& \multicolumn{2}{c}{mIoU ↑} & \multicolumn{2}{c}{Pixel Acc. ↑}
& \multicolumn{2}{c}{RMSE ↓} & \multicolumn{2}{c}{Angle ↓} \\
\cmidrule(lr){2-3}\cmidrule(lr){4-5}\cmidrule(lr){6-7}\cmidrule(lr){8-9}
\textbf{Model}
& {Base} & {Ours} & {Base} & {Ours}
& {Base} & {Ours} & {Base} & {Ours} \\
\midrule
Small & 70.1 & \bfseries 72.7 & 93.6 & \bfseries 94.6 & 0.422 & \bfseries 0.421 & 42.8 & \bfseries 39.6 \\
Base  & 73.3 & \bfseries 75.7 & 94.5 & \bfseries 95.5 & 0.388 & \bfseries 0.380 & 41.3 & \bfseries 39.4 \\
Large & 74.0 & \bfseries 76.5 & 94.7 & \bfseries 95.7 & 0.381 & \bfseries 0.351 & 42.0 & \bfseries 38.6 \\
Huge+ & 73.2 & \bfseries 75.9 & 94.4 & \bfseries 95.6 & 0.407 & \bfseries 0.367 & 42.8 & \bfseries 38.2 \\
\bottomrule
\end{tabular}
\end{table}

\subsection{Evaluation of computational cost}
\begin{figure}
    \centering
    \includegraphics[width=\columnwidth]{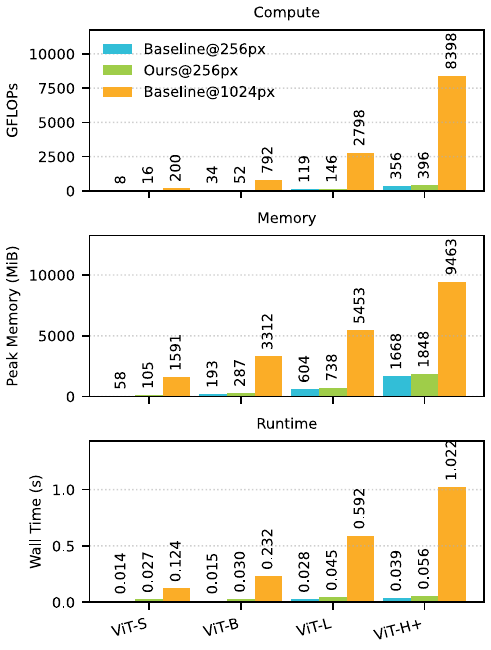}
    \caption{We compare the computational cost of generating dense features of size 64x64 for a single image using the DINOv3 baseline (1024 pixels) and the proposed method (256 pixels). Blue bars additionally show the computational cost of the baseline model when computing a 16x16 feature map. Runtime is measured on an NVIDIA A100.}
    \label{fig:complexity}
\end{figure}

If input images are available in high resolution, one can compute very dense feature maps using baseline models. We demonstrate that the proposed method has merit in that scenario by analyzing the computational cost of generating feature maps of size 64x64 in \cref{fig:complexity}. By avoiding quadratic complexity, we observe huge savings in terms of FLOPS, run time and memory, allowing us to generate features using the Huge+ model on a laptop with 4GBs of VRAM. As downstream applications often rely on large amounts of inference model calls, BRIXEL promises substantial resource savings for practitioners. In terms of model size, BRIXEL increases the parameter count by an average of $10\%$ across ViT sizes.

\subsection{How dense can the student get?}
\begin{figure*}
\centering
    \includegraphics[width=\textwidth]{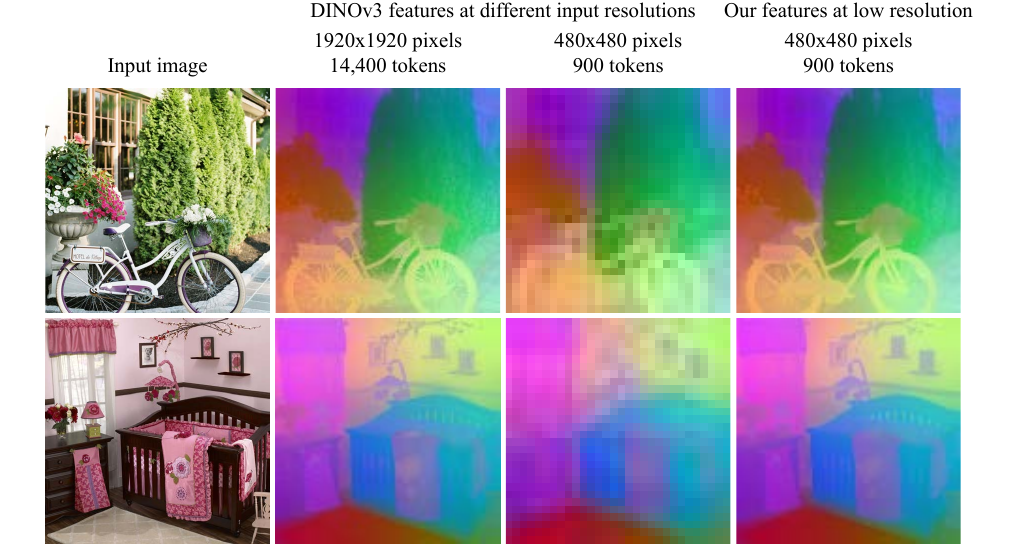}
    \caption{Feature maps of the ViT-B model fine-tuned and evaluated at an image size of 480x480. Best viewed on screen using zoom.}
    \label{fig:finetuned_panel}
\end{figure*}
\begin{table}[t]
\centering
\caption{Performance of both scene-centric and object-centric semantic segmentation of the ViT-Base student fine-tuned at higher resolution and evaluated with 512 pixels per side. Fine-tuning refers only to the knowledge distillation; training on labeled benchmarks is still restricted to linear probes. 'Baseline' refers again to the original DINOv3 model, now also evaluated at an image size of 512.}
\label{tab:higher-resolution}
\begin{tabular}{l
                S[table-format=2.3] S[table-format=2.3]
                S[table-format=2.2] S[table-format=2.2]}
\toprule
& \multicolumn{2}{c}{mIoU ↑} & \multicolumn{2}{c}{Pixel Accuracy ↑} \\
\cmidrule(lr){2-3}\cmidrule(lr){4-5}
\textbf{Dataset} & {Baseline} & {Ours} & {Baseline} & {Ours} \\
\midrule
ADE20k   & 51.4 & \bfseries \textbf{52.2} & 82.7 & \bfseries \textbf{83.2} \\
Cityscapes  & 70.0 & \bfseries \textbf{70.3} & 93.8 & \bfseries \textbf{94.3} \\
PASCAL VOC   & 75.6 & \bfseries \textbf{76.5} & 95.5 & \bfseries \textbf{95.8} \\
\bottomrule
\end{tabular}
\end{table}

So far, we have trained and evaluated the student at a resolution of 256, the common input size for ViTs, which yields the original 16x16 words that an image is worth \cite{dosovitskiyImageWorth16x162021}. At this resolution, we have shown that the student outperforms the baseline DINOv3 model by a considerable margin. Next, we examine whether performance on dense tasks can be improved at higher resolution as well. To this end, we fine-tune the ViT-B model on images with 480 pixels per side, meaning that the teacher network operates at a resolution of 1920, corresponding to $120^2 = 14400$ tokens. Due to the high resolution of the teacher network, fine-tuning is substantially more expensive than the original training pipeline. As our NVIDIA A100s can only fit one teacher sample at a time, we parallelize training across 8 A100s. These differences aside, the fine-tuning procedure adheres to the same layout as the original training outlined above.

\begin{figure}
    \centering
    \includegraphics[width=\linewidth]{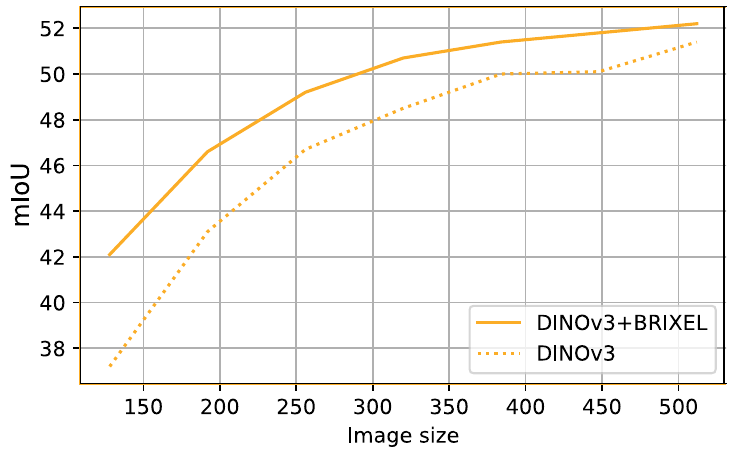}
    \caption{Segmentation performance of the finetuned BRIXEL model evaluated at a variety of image sizes.}
    \label{fig:sizewise_main}
\end{figure}

We show qualitative results in \cref{fig:finetuned_panel}. As before, we observe that the student is able to produce highly faithful copies of the teacher's feature maps. To evaluate the fine-tuned model quantitatively, we rerun the experiments on ADE20k, Cityscapes and PASCAL VOC analogously to the previous sections. As these benchmarks are often tackled at a resolution of 512, we also evaluate at this image size despite it being slightly larger than the training size. \cref{tab:higher-resolution}
displays the resulting metrics. Again, the student consistently outperforms the teacher on all three datasets and both metrics when keeping the input resolution fixed. This demonstrates that BRIXEL not only has strong merit for budget-friendly computing at low resolution, but also when aiming for extremely fine-grained feature maps. 
We also examine the ability of the student to generalize to other image sizes not observed during training. \cref{fig:sizewise_main} shows that the finetuned BRIXEL model outperforms the DINOv3 baseline on ADE20k at all image sizes between 128 and 512, showing that BRIXEL is applicable to downstream tasks at any image size. 

\subsection{Loss ablations}
The full loss used for training BRIXEL consists of a weighted sum of the terms $L_1$, $L_\text{edge}$ and $L_\text{spectral}$. 
To study the influence of each loss term, we train multiple BRIXEL models from scratch using the same experimental setup as before, but considering different subsets of these terms. To probe performance, we then rerun the segmentation experiments on ADE20k, as described above, for each model. \cref{tab:ablation} displays the reduction of mIoU of each model compared to the one using the full loss function. We observe that all loss terms contribute to BRIXEL's performance, with $L_1$ and $L_\text{edge}$ being the most important.

\begin{table}[t]
\setlength{\tabcolsep}{3.5pt}
\centering
\caption{Ablation results for the loss function used for training the BRIXEL model. We retrain the BRIXEL model with different subsets of the loss terms, and compute the reduction of mIoU of a lightweight segmentation probe on ADE20k compared to the full loss.}
\label{tab:ablation}
\begin{tabular}{lccc}
\toprule
\textbf{Dataset} & $L_1$+Edge & $L_1$+Spectral & $L_1$ only \\
\midrule
ADE20k & -0.3 & -5.0 & -0.6 \\
Cityscapes & -0.0 & -4.1 & -0.4 \\
Pascal & -0.1 & -6.5 & -0.4 \\
\bottomrule
\end{tabular}
\end{table}

\subsection{Beyond DINOv3 backbones}
\label{subsec:beyond_dino}
\begin{figure}
    \centering
    \includegraphics[width=1.0\linewidth]{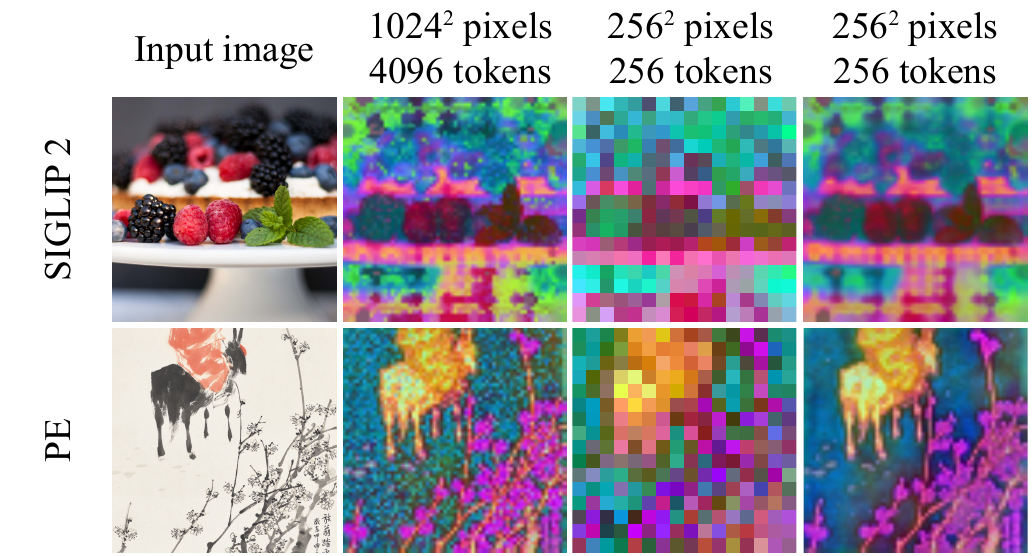}
    \caption{Qualitative results for SIGLIP 2 and Perception Encoder.}
    \label{fig:beyond_qualitative}
\end{figure}

% \begin{figure*}
% \centering
%     \includegraphics[width=\textwidth]{Figures/siglip_qualitative.pdf}
%     \caption{Feature maps computed using SigLIP 2 as a backbone instead of DINOv3.}
%     \label{fig:siglip}
% \end{figure*}
This work was originally inspired by the size-agnostic pretraining of DINOv3. However, given the convincing results for this model family, we wondered how well BRIXEL can reconstruct feature maps of other foundation models. To this end, we trained the student-teacher setup as described above for the base variant of the recent SigLIP \cite{tschannenSigLIPMultilingualVisionLanguage2025} and Perception Encoder (P. ENC.) \cite{bolyaPerceptionEncoderBest2025} models. For Perception encoder, we study the 'spatial' variant as it yields the cleanest dense features. We train BRIXEL for these models using the same setup as for DINOv3. Qualitative results are shown in \cref{fig:beyond_qualitative} and indicate that BRIXEL produces faithful high-resolution feature maps for these model families as well.

To study quantitative performance, we rerun all downstream performance probes described above for SIGLIP 2 and Perception Encoder. These results are shown in \cref{tab:beyond-backbones}. We observe an increase in performance comparable to that for DINOv3, showing that BRIXEL does generalize well across different pretraining recipes.

\begin{table}[t]
\centering
\small
\setlength{\tabcolsep}{3.5pt}
\caption{Performance of BRIXEL paired with other dense-feature extractors.}
\label{tab:beyond-backbones}

% -------- Segmentation (ADE20k / Cityscapes / PASCAL) --------
\begin{tabular}{ll
S[table-format=2.1] S[table-format=2.1]
S[table-format=2.1] S[table-format=2.1]}
\toprule
& & \multicolumn{2}{c}{mIoU ↑} & \multicolumn{2}{c}{Pixel Accuracy ↑} \\
\cmidrule(lr){3-4}\cmidrule(lr){5-6}
\textbf{Dataset} & \textbf{Model} & {Baseline} & {Ours} & {Baseline} & {Ours} \\
\midrule

\multirow{3}{*}{ADE20k}
 & DINOv3   & 46.7 & \bfseries 49.2 & 80.5 & \bfseries 82.0 \\
 & SIGLIP 2 & 36.6 & \bfseries 40.8 & 72.0 & \bfseries 76.0 \\
 & P. ENC.  & 34.0 & \bfseries 37.8 & 73.3 & \bfseries 76.3 \\
\midrule

\multirow{3}{*}{Cityscapes}
 & DINOv3   & 61.1 & \bfseries 64.4 & 91.6 & \bfseries 93.0 \\
 & SIGLIP 2 & 44.2 & \bfseries 50.0 & 88.1 & \bfseries 90.9 \\
 & P. ENC.  & 45.7 & \bfseries 52.0 & 90.0 & \bfseries 92.4 \\
\midrule

\multirow{3}{*}{PASCAL}
 & DINOv3   & 73.3 & \bfseries 75.7 & 94.5 & \bfseries 95.5 \\
 & SIGLIP 2 & 61.9 & \bfseries 67.1 & 90.1 & \bfseries 92.2 \\
 & P. ENC.  & 61.0 & \bfseries 65.4 & 90.8 & \bfseries 92.8 \\
\bottomrule
\end{tabular}

\vspace{6pt}

% -------- Depth / Normals (NYU / NYU-2) --------
% -------- Depth / Normals (NYU / NYU-2) --------
\begin{subtable}{\linewidth}
\centering
\begin{tabular}{ll
                S[table-format=1.3] S[table-format=1.3]
                S[table-format=2.2] S[table-format=2.2]}
\toprule
& & \multicolumn{2}{c}{RMSE ↓} & \multicolumn{2}{c}{Angular Error ↓} \\
\cmidrule(lr){3-4}\cmidrule(lr){5-6}
\textbf{Dataset} & \textbf{Model} & {Baseline} & {Ours} & {Baseline} & {Ours} \\
\midrule

\multirow{3}{*}{NYU}
 & DINOv3   & 0.354 & \bfseries 0.346 & {--} & \bfseries {--} \\
 & SIGLIP 2 & 0.613 & \bfseries 0.567 & {--} & \bfseries {--} \\
 & P. ENC.  & 0.394 & \bfseries 0.381 & {--} & \bfseries {--} \\
\midrule
\multirow{3}{*}{NAVI}
 & DINOv3   & 0.388 & \bfseries 0.380 & 41.31 & \bfseries 39.35 \\
 & SIGLIP 2 & 0.506 & \bfseries 0.489 & 49.37 & \bfseries 48.62 \\
 & P. ENC.  & 0.469 & \bfseries 0.440 & 48.42 & \bfseries 44.87 \\
\bottomrule

\end{tabular}
\end{subtable}

\end{table}

\section{Related work}
Our work builds on previous efforts to produce task-specific pixel-wise annotations based on low-resolution ViT features for tasks like depth estimation and segmentation \cite{chengPixelClassificationNot2021, chenVisionTransformerAdapter2022, chengMaskedattentionMaskTransformer2022, yangDepthAnythingV22024, raviSAMSegmentAnything2024, simeoniDINOv32025}, object detection \cite{simeoniLocalizingObjectsSelfsupervised2021a, wangTokenCutSegmentingObjects2023} and localization \cite{zhouExtractFreeDense2022, luoBRAINMAPPINGDENSE2025}. Our approach falls in line with methods aimed towards improving the quality of general dense feature maps \emph{before} applying them to downstream tasks. To this end, previous work has explored alternative global attention mechanisms \cite{zhouExtractFreeDense2022}, smoothing out high-norm patch tokens tokens \cite{wangSINDERRepairingSingular2024} as well as enforcing spatial consistency \cite{luoBRAINMAPPINGDENSE2025}. Highly relevant is also recent literature on register tokens \cite{darcetVisionTransformersNeed2024}, which aim to improve the quality of dense feature maps by relaxing constraints on how global information is extracted from patch features \cite{chenVisionTransformersSelfDistilled2025, jiangVisionTransformersDont2025, lappeRegisterCLSTokens2025}. Conceptually, BRIXEL differs by not attempting to modify the feature space itself, but rather keeping the features as aligned as possible with the high-resolution teacher. Knowledge distillation \cite{hintonDistillingKnowledgeNeural2015}, which BRIXEL is based on, has been explored extensively in recent years, particularly in the context of vision foundation models \cite{caronEmergingPropertiesSelfSupervised2021a, oquabDINOv2LearningRobust2023, touvronTrainingDataefficientImage2021}. In fact, all models from the DINOv3 family considered in this work were trained by distilling knowledge from a very large teacher network \cite{simeoniDINOv32025}.

\section{Conclusion}
We have shown that, using a simple self-distillation strategy, we can faithfully increase the resolution of dense feature maps extracted by recent vision foundation models. The resulting features consistently outperform the baselines across a variety of tasks and benchmarks. Our findings can be used to both improve performance when the input resolution is fixed, as well as to generate higher-resolution maps when high-quality input images are not available, or when the computational budget is limited. 

\paragraph{Limitations and future work.} 
This project investigates the dense feature maps at the output of the model. As some work suggests also using intermediate features for spatial tasks, future efforts could incorporate the reconstruction of higher-resolution features at multiple stages. Further, the mechanisms through which the ViTs studied in this work facilitate the upsampling procedure are still unclear and should be investigated in future work. One promising avenue might be a systematic study of how much fine-grained spatial information is encoded \emph{within} each transformer token.

\paragraph{Acknowledgements.}
AL and MG are supported by ERC-SyG 856495. MG is supported by HFSP RGP0036/2016, BMBF
FKZ 01GQ1704. The authors thank the International Max Planck Research School for Intelligent Systems (IMPRS-IS)
for supporting Alexander Lappe.

\FloatBarrier

{
    \small
    \bibliographystyle{ieeenat_fullname}
    \bibliography{RegisterModels}
}

% \clearpage
% \appendix
% \section{Appendix}
% \FloatBarrier
% % Maybe mention the hann windowing and cite {harrisUseWindowsHarmonic1978}

% % Auch noch das gleiche Bild für alle Modelle für Modellvergleich

% \begin{figure*}
% \centering
%     \includegraphics[width=\textwidth]{Figures/sizewise.pdf}
%     \caption{We evaluate the fine-tuned ViT-B BRIXEL model on semantic segmentation on ADE20k at a variety of input image sizes. BRIXEL outperforms the DINOv3 baseline at all image sizes, showing that the method generalizes well beyond the training image size.}
%     \label{fig:sizewise}
% \end{figure*}

% \begin{figure*}
% \centering
%     \includegraphics[width=\textwidth]{Figures/sizewise_qualitative.pdf}
%     \caption{Feature maps of the fine-tuned ViT-B model for different input sizes. Different RGB maps across image sizes are due to the PCA being recomputed for each image size.}
%     \label{fig:sizewise_qualitative}
% \end{figure*}

\input{sec/X_suppl}

\end{document}

%% file: sec/X_suppl.tex
\clearpage
\setcounter{page}{1}
\maketitlesupplementary

\onecolumn
\begin{figure}
% \onecolumn
\centering
    \includegraphics[width=\textwidth]{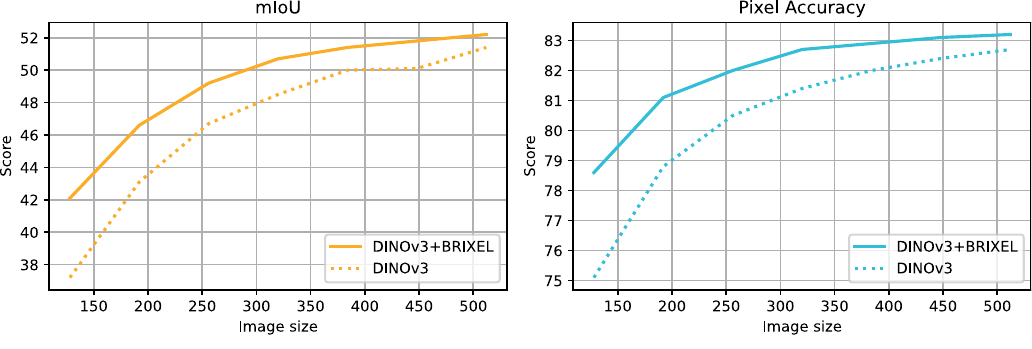}
    \caption{We evaluate the fine-tuned ViT-B BRIXEL model on semantic segmentation on ADE20k at a variety of input image sizes. BRIXEL outperforms the DINOv3 baseline at all image sizes, showing that the method generalizes well beyond the training image size.}
    \label{fig:sizewise}
\end{figure}

\begin{figure}
\centering
    \includegraphics[width=\textwidth]{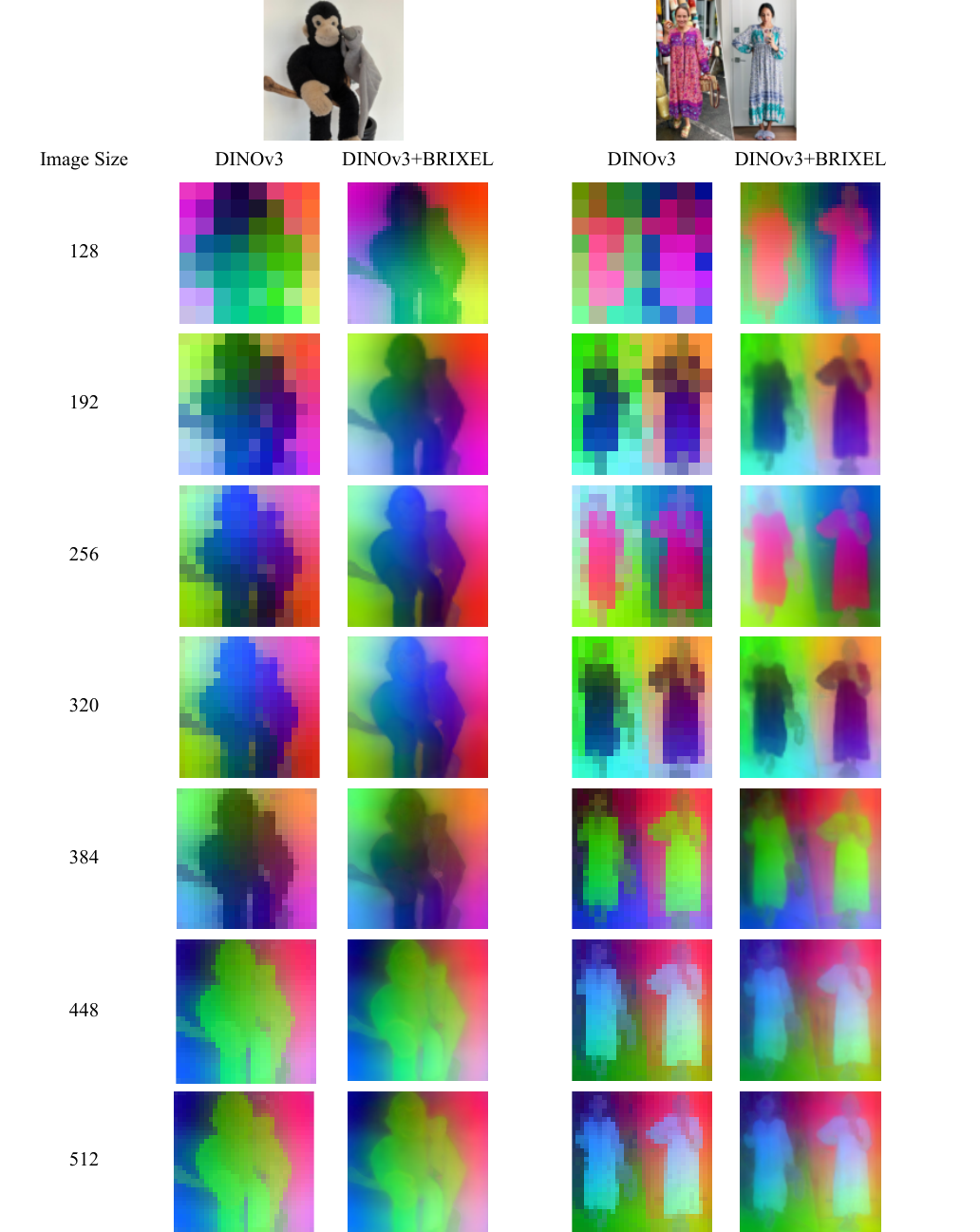}
    \caption{Feature maps of the fine-tuned ViT-B model for different input sizes. Different RGB maps across image sizes are due to the PCA being recomputed for each image size.}
    \label{fig:sizewise_qualitative}
\end{figure}